\documentclass{article} 
\usepackage{tencent_ailab_tech_report}
\usepackage[colorlinks = true,
            linkcolor = blue,
            urlcolor  = blue,
            citecolor = blue,
            anchorcolor = blue]{hyperref}

\usepackage{graphicx}
\usepackage{caption}
\usepackage{subcaption}
\usepackage{booktabs} 

\usepackage[colorinlistoftodos]{todonotes}

\usepackage{listings}
\usepackage{xcolor}
\usepackage{natbib}
\usepackage{bbding}
\lstdefinelanguage{json}{
    basicstyle=\ttfamily\small,
    numbers=left,
    numberstyle=\tiny,
    stepnumber=1,
    numbersep=5pt,
    showstringspaces=false,
    breaklines=true,
    frame=lines,
    backgroundcolor=\color{gray!10},
    literate=
     *{0}{{{\color{blue}0}}}{1}
      {1}{{{\color{blue}1}}}{1}
      {2}{{{\color{blue}2}}}{1}
      {3}{{{\color{blue}3}}}{1}
      {4}{{{\color{blue}4}}}{1}
      {5}{{{\color{blue}5}}}{1}
      {6}{{{\color{blue}6}}}{1}
      {7}{{{\color{blue}7}}}{1}
      {8}{{{\color{blue}8}}}{1}
      {9}{{{\color{blue}9}}}{1}
      {:}{{{\color{red}:}}}{1}
      {,}{{{\color{red},}}}{1}
      {"}{{{\color{orange}"}}}{1}
      {[}{{{\color{green}[}}}{1}
      {]}{{{\color{green}]}}}{1}
      {\{}{{{\color{green}\{}}}{1}
      {\}}{{{\color{green}\}}}}{1}
}

\usepackage{microtype}
\usepackage{hyperref}
\usepackage{url}
\usepackage{booktabs}
\usepackage{enumitem}
\usepackage{multicol}
\usepackage{multirow}
\usepackage{CJKutf8}
\usepackage{amsmath}
\usepackage{siunitx}
\usepackage{floatflt}
\usepackage{wrapfig}
\usepackage{graphicx}
\usepackage{booktabs}
\usepackage{wrapfig}
\usepackage{authblk}
\usepackage{lipsum}
\usepackage{dsfont}
\usepackage{bbm}
\usepackage{microtype}
\usepackage{graphicx}
\usepackage{subcaption}
\usepackage{multirow}
\usepackage{booktabs} 
\usepackage{pifont}  
\usepackage{graphicx}  
\usepackage{subcaption} 
\usepackage{colortbl}
\usepackage{hyperref}

\usepackage[ruled,vlined,linesnumbered,noend]{algorithm2e}
\usepackage{amsmath,amssymb,xcolor,setspace}

\SetKwInput{KwIn}{Require}
\SetKwInput{KwOut}{Return}
\SetKwComment{Comment}{$\triangleright$\ }{}
\SetKw{KwTo}{to}
\SetArgSty{textnormal}
\SetAlCapNameFnt{\normalsize\bfseries}
\SetAlCapFnt{\normalsize\bfseries}
\usepackage{marvosym}
\usepackage[most]{tcolorbox}
\newtcolorbox{alprompt}[1]{
        boxrule = 1pt,
        fontupper = \small\tt,
        fonttitle = \bf\color{black},
        arc = 2pt,
        rounded corners,
        colframe = black,
        colbacktitle = white!97!yellow,
        colback = white!97!yellow,
        title = #1,
}
\definecolor{darkgreen}{rgb}{0.0, 0.5, 0.0}
\definecolor{darkgray}{gray}{0.4}
\definecolor{maroon}{rgb}{0.5, 0.0, 0.0}
\definecolor{navy}{rgb}{0.0, 0.0, 0.5}
\definecolor{teal}{rgb}{0.0, 0.5, 0.5}

\definecolor{deepblue}{RGB}{41, 128, 185}

\definecolor{mylightgreen}{RGB}{144,238,144}
\definecolor{mylightblue}{RGB}{173,216,230}

\definecolor{outerboxcolor}{gray}{0.90} 
\definecolor{innerboxcolor}{rgb}{1,1,1}

\definecolor{nred}{RGB}{196, 38, 11}
\definecolor{ngreen}{RGB}{18, 141, 21}
\definecolor{nblue}{RGB}{41, 52, 190}

\usepackage{listings}

\usepackage{array}
\usepackage{amsmath}
\usepackage{mathtools}
\usepackage{amsthm}
\usepackage{arydshln}
\usepackage[capitalize,noabbrev]{cleveref}
\usepackage{adjustbox} 
\usepackage{enumitem}
\usepackage{longtable}
\usepackage{tcolorbox}
\tcbuselibrary{skins}

\theoremstyle{plain}

\theoremstyle{definition}

\theoremstyle{remark}

\makeatletter
\renewcommand{\@fnsymbol}[1]{\ensuremath{\ifcase#1\or \dag\or \ddag\or \S\or \P\or \|\or **\fi}}
\makeatother

\newcommand{\method}{\textsc{R-Few}}

\colmfinalcopy

\title{
\begin{center}
Guided Self-Evolving LLMs with Minimal Human Supervision
\end{center}
}

\author[]{Wenhao Yu$^1$, Zhenwen Liang$^1$, Chengsong Huang$^2$, Kishan Panaganti$^1$, \\
\vspace{-1.0em}\textbf{Tianqing Fang$^1$, Haitao Mi$^1$, Dong Yu$^1$}
\\
$^1$Tencent AI Lab in Seattle,
$^2$Washington University in St. Louis
\\
\texttt{wenhaowyu@global.tencent.com}
}

\begin{document}

\maketitle

\begin{abstract}

AI self-evolution has long been envisioned as a path toward superintelligence, where models autonomously acquire, refine, and internalize knowledge from their own learning experiences. Yet in practice, unguided self-evolving systems often plateau quickly or even degrade as training progresses. These failures arise from issues such as concept drift, diversity collapse, and mis-evolution, as models reinforce their own biases and converge toward low-entropy behaviors.
To enable models to self-evolve in a stable and controllable manner while minimizing reliance on human supervision, we introduce \textsc{R-Few}, a guided Self-Play Challenger–Solver framework that incorporates lightweight human oversight through in-context grounding and mixed training. At each iteration, the Challenger samples a small set of human-labeled examples to guide synthetic question generation, while the Solver jointly trains on human and synthetic examples under an online, difficulty-based curriculum.
Across math and general reasoning benchmarks, R-Few achieves consistent and iterative improvements. For example, Qwen3-8B-Base improves by +3.0 points over R-Zero on math tasks and achieves performance on par with General-Reasoner, despite the latter being trained on 20 times more human data. Ablation studies confirm the complementary contributions of grounded challenger training and curriculum-based solver training, and further analysis shows that R-Few mitigates drift, yielding more stable and controllable co-evolutionary dynamics.
\end{abstract}

\begin{figure}[h]
    \centering
    \vspace{-0.1in}
    \includegraphics[width=0.95\linewidth]{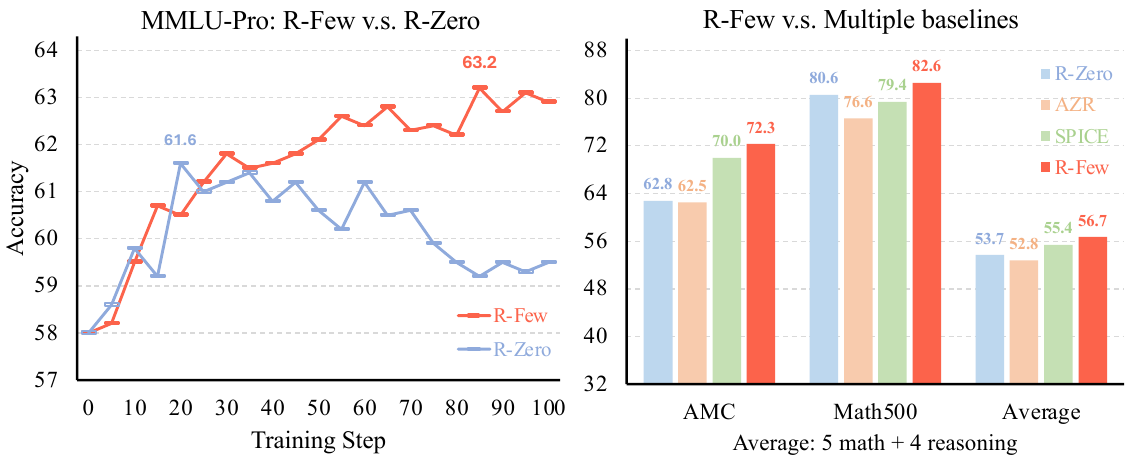}
    \vspace{-0.1in}
    \caption{\method{} delays the performance plateau seen in R-Zero and achieves higher performance. After training, it outperforms baselines multiple benchmarks, showing more stable self-evolution.}
    \vspace{-0.1in}
\end{figure}
\section{Introduction}

A long-standing vision in AI research is the development of systems that can self-evolve. It refers to a model that can autonomously acquire, refine, and internalize knowledge from its own generated experiences, thereby achieving continuous self-improvement without relying on large human-labeled datasets~\citep{Tao2024ASO}.
To scale these capabilities without human
supervision, self-play offers a promising paradigm~\citep{silver2017mastering,sukhbaatar2017intrinsic}, where models
improve by competing against themselves and generating automatic feedback through competition.

However, when applied to language tasks, self-play remains surprisingly brittle. Despite early gains, methods such as Absolute Zero and R-Zero often plateau quickly, and their performance may even deteriorate as training progresses.~\citep{zhao2025absolute,huang2025r,chen2025self,kuba2025language,he2025visplay}.
Two key challenges underlie this phenomenon:
(i) \textit{Concept Drift}: Without external intervention or guidance, models tend to reinforce their own knowledge bias. Over repeated self-play iterations, feedback loops amplify spurious correlations or biased reasoning patterns, drifting away from factual correctness and semantic validity~\citep{liu2025spice}.
(ii) \textit{Diversity Collapse}: Since the underlying knowledge of the model is fixed, its self-generated challenges tend to converge toward familiar and low-entropy regions of the task space. As the diversity of generated tasks diminishes, exploration and reasoning novelty stagnate~\citep{liang2025beyond,morris2025much}.

Together, these problems make it difficult for language models to evolve in a stable and controllable manner. 
To overcome this challenge, we need a framework that minimizes reliance on human annotations while its evolution should remain guided toward human-intended goals.
In this work, we propose \method{}, a self-evolving LLM framework that integrates limited human supervision into the self-play process. Unlike prior approaches that require large-scale labeled corpora or extensive human feedback, \method{} leverages only a small pool of high-quality ``anchor'' data -- specifically, 1\% to 5\% data sampled from WebInstruct~\citep{Ma2025GeneralReasonerAL}, a open-source large-scale dataset mined from the web. During training, the Challenger dynamically samples zero to five anchor examples as in-context example to guide synthetic question generation. When zero examples are sampled, \method{} preserves the ability for open-ended exploration; when few-shot examples are included, they inject semantic anchors that regularize the Challenger’s generation.
To further enhance the learning dynamics, we introduce an online curriculum mechanism for the Solver. Instead of indiscriminately training on all generated data, \method{} first rolls out multiple solving trajectories, ranks questions by the Solver’s uncertainty, and selects mid-uncertain samples for the next iteration training. This ranking-based curriculum ensures that the Solver focuses on the zone of proximal development. The same selection principle is applied to both synthetic and human anchor data, so it can merge human-curated and self-generated question into a unified training stream for Solver.

Our experiments demonstrate that \method{} consistently and iteratively improving the reasoning abilities. For example, Qwen3-8B-Base model's average score on math benchmarks increased by a significant +3.0 points than R-Zero, achieving on par performance with General-Reasoner~\citep{Ma2025GeneralReasonerAL}, the model trained on the full WebInstruct dataset (232k). 
In addition, we provide an in-depth analysis that validates our framework's components, and characterizes the co-evolutionary dynamics to identify both strengths and limitations, offering insights for future research.

\section{Preliminaries: Self-Play For Data-Free Training}
\label{sec:preli}

Inspired by the success of self-play mechanisms in games, exemplified by AlphaZero~\citep{silver2017mastering,Silver2017MasteringTG}, recent research have begun extending this paradigm to large language models (LLMs), aiming to create self-evolving systems that can continually refine their reasoning abilities without curated human data~\citep{chen2024self}.
As base model capabilities have risen and reinforcement learning has advanced, a wave of ``self-evolving + RL'' training schemes has emerged~\citep{zhao2025absolute,huang2025r,chen2025self,liu2025spice}. These approaches show that models can autonomously construct their own curricula -- posing challenges, solving them, and improving through iteration -- making self-improving intelligence an increasingly attainable reality.

\subsection{Unified View of Self-Play Objectives and Reward Design}
\label{sec:prelim:unified}

Let \( Q_\theta \) denote the \textbf{Challenger} (aka. Proposer, Questioner) and \( S_\phi \) the \textbf{Solver} (aka. Reasoner), both initialized from the same pretrained LLM \( M_0 \).
At each iteration \( t \), the Challenger samples a question or task $q_t \sim Q_\theta(\cdot \mid \mathcal{H}_t)$,
where \( \mathcal{H}_t \) represents the historical context (e.g., previous questions and responses, document corpus), and the Solver attempts to generate an answer $a_t \sim S_\phi(\cdot \mid q_t)$.
A verifier or self-consistency mechanism then evaluates outputs and produces a scalar reward $r_t = \mathcal{R}(q_t, a_t; S_\phi)$,
which guides the policy updates of both agents via reinforcement learning, typically using Group Relative Policy Optimization (GRPO)~\citep{shao2024deepseekmath}:
\begin{align}
    \theta &\leftarrow \theta + \eta_\theta \nabla_\theta \, \mathbb{E}_{q_t \sim Q_\theta} [\mathcal{R}_{\text{chal}}(q_t)], \\
    \phi &\leftarrow \phi + \eta_\phi \nabla_\phi \, \mathbb{E}_{a_t \sim S_\phi(q_t)} [\mathcal{R}_{\text{sol}}(q_t, a_t)].
\end{align}

\paragraph{Unified objective.}
The overall bi-level optimization can be expressed as
\begin{equation}
\label{eq:bilevel}
\max_{\theta}\ \mathbb{E}_{q\sim Q_\theta}\big[ \mathcal{R}_{\text{chal}}(q;\phi) \big]
\quad\text{s.t.}\quad
\phi \leftarrow \arg\max_{\phi}\ \mathbb{E}_{q\sim Q_\theta,\,a\sim S_\phi(q)} \big[ \mathcal{R}_{\text{sol}}(q,a) \big].
\end{equation}
This formulation unifies language self-play methods under a single reinforcement learning objective, where the Challenger adaptively generates a curriculum of problems and the Solver improves its reasoning skills in response.

\paragraph{Reward design across frameworks.}
Different self-play frameworks instantiate distinct choices of the Challenger and Solver rewards \( (\mathcal{R}_{\text{chal}}, \mathcal{R}_{\text{sol}}) \), each encoding a specific learning bias:

\begin{itemize}[leftmargin=*]
    \vspace{-0.1in}
    \item \textbf{Absolute Zero}~\citep{zhao2025absolute}: Operates in a fully zero-data regime, using a linear difficulty shaping function to maintain stability:
    \[
        \mathcal{R}_{\text{chal}}^{\text{AbsZero}}(q) 
= \left( 1 - \widehat{p}_{\text{succ}}(q) \right) \cdot \mathbb{I}\left( \widehat{p}_{\text{succ}}(q) \neq 0 \right)
,
        \qquad
        \mathcal{R}_{\text{sol}}^{\text{AbsZero}}(q,a)
        = \mathbb{I}\{a = y(q)\},,
    \]
    where \( f \) peaks near \( \widehat{p}_{\text{succ}} = 0 \)
    \vspace{-0.05in}
    \item \textbf{R-Zero}~\citep{huang2025r}: Employs a purely uncertainty-driven curriculum with verifiable pseudo-labels:
    \[
        \mathcal{R}_{\text{chal}}^{\text{R-Zero}}(q)
        = 1 - 2\big|\widehat{p}_{\text{succ}}(q) - \tfrac{1}{2}\big|
        - \lambda_{\text{rep}}\mathrm{RepPenalty}(q),
        \qquad
        \mathcal{R}_{\text{sol}}^{\text{R-Zero}}(q,a)
        = \mathbb{I}\{a = \tilde{y}(q)\},
    \]
    where \( \tilde{y}(q) \) is the majority-vote pseudo-answer and \( \mathrm{RepPenalty} \) discourages duplicate questions.
    \vspace{-0.05in}

    \item \textbf{Self-Questioning LMs (SQLM)}~\citep{chen2025self}: The Challenger optimizes for both \textit{answerability} and \textit{informativeness} by targeting a difficulty sweet spot based on the success rate statistics:
    \[
        \mathcal{R}_{\text{chal}}^{\text{SQLM}}(q) = \left( 1 - \widehat{p}_{\text{succ}}(q) \right) \cdot \mathbb{I}\left( 0 < p_{\text{succ}}(q) < \widehat{p}_{\text{succ}}(q) \right),
        \qquad
        \mathcal{R}_{\text{sol}}^{\text{SQLM}}(q,a) = \mathbb{I}\{a = \tilde{y}(q)\},
    \]
    where \(p_{\text{succ}}(q)\) denotes the pass rate, and \(\tilde{y}(q)\) represents the majority-vote pseudo-answer.
    \vspace{-0.05in}
    \item \textbf{SPICE}~\citep{liu2025spice}: Uses a \emph{corpus-guided} framework where the Challenger is rewarded for generating questions with a specific response variance (indicating appropriate difficulty), while the Solver is supervised by the source text:
    \[
        \mathcal{R}_{\text{chal}}^{\text{SPICE}}(q)
        = \mathbb{I}(q \text{ is valid}) \cdot \exp \left( -\frac{(\text{Var}(\{\hat{y}_1, \dots, \hat{y}_K\}) - 0.25)^2}{2 \cdot 0.01} \right),
        \quad
        \mathcal{R}_{\text{sol}}^{\text{SPICE}}(q,a)
        = \mathbb{I}\{a = y(q)\},
    \]
    where \(y(q)\) is the ground-truth answer extracted directly from the reference document, and \(\{\hat{y}_k\}\) represents the set of sampled responses from peer models.
\end{itemize}


\begin{figure}[th]
    \centering
    \includegraphics[width=\linewidth]{figures/rfew-framework.pdf}
    \vspace{-0.2in}
    \caption{\textbf{The math examples shown in the figure are not real data}, but are included for demonstration to aid understanding. The figure provides an overview of our \method{} framework. The Challenger is incentivized to generate moderately (``medium'') uncertain questions that lie at the edge of the Solver’s current abilities; the Solver is rewarded for solving increasingly challenging tasks -- sourced from both humans and the Challenger -- via curriculum-based selection.}
    \vspace{-0.1in}
\label{fig:method}
\end{figure}

\section{Method}
\label{sec:method}
\subsection{Overview}

Building upon the data-free self-play foundation established, we introduce \textsc{\method{}}, a self-evolving framework that integrates \emph{limited human supervision} 
to stabilize and accelerate reasoning evolution. 
\textsc{\method{}} presents two key innovations:
(1) a \textbf{few-shot grounded Challenger} that anchors synthetic task generation to small human ``anchor'' data, and 
(2) an \textbf{online curriculum Solver} that adaptively selects mid-uncertain samples from both synthetic and human sources.

\subsection{Few-Shot Grounded Challenger}
\label{sec:method:challenger}

The Challenger \( Q_\theta \) in \textsc{\method{}} generates synthetic questions conditioned on a small pool of high-quality 
human anchor examples \( \mathcal{D}_{\text{H}} = \{(x_i, y_i)\}_{i=1}^{N_{\text{H}}} \). 
At each rollout step, we randomly sample \( k \in \{0,1,\dots,5\} \) demonstrations as in-context examples:
\begin{equation}
    \mathcal{C}_t = \mathrm{Sample}_k(\mathcal{D}_{\text{H}}), 
    \qquad
    q_t \sim Q_\theta(\cdot \mid \mathcal{C}_t, \mathcal{H}_t).
\end{equation}
When \( k = 0 \), \textsc{\method{}} degenerates to the data-free setting of \textsc{R-Zero}, 
preserving open-ended self-evolution. For \( k>0 \), the Challenger is softly guided toward 
human-consistent reasoning trajectories while retaining generative diversity.

\begin{equation}
\label{eq:rfew-chal}
\mathcal{R}_{\text{chal}}^{\text{\method{}}}(q_t) 
= \underbrace{1 - 2\big|\widehat{p}_{\text{succ}}(q_t) - \tfrac{1}{2}\big|}_{\text{difficulty shaping}}
-\lambda_{\text{rep}}\,\mathrm{RepPenalty}(q_t).
\end{equation}
Here, \(\mathrm{Align}(q_t, \mathcal{D}_{\text{H}})\) measures the semantic or structural proximity between the 
generated question \( q_t \) and the human anchor distribution via cosine similarity in embedding space.
This term encourages the Challenger to explore \emph{in the vicinity} of real human data, 
preventing semantic drift while maintaining autonomous difficulty progression.

The Challenger parameters are updated via GRPO as:
\begin{equation}
    \theta \leftarrow \theta + \eta_\theta \nabla_\theta 
    \mathbb{E}_{q_t \sim Q_\theta} [\mathcal{R}_{\text{chal}}^{\text{\method{}}}(q_t)].
\end{equation}

Besides, there is another trick we used during challenger training. Because the base model sometimes struggles to follow instructions when the prompt gets longer, we added a quick warm-up stage via supervised fine-tuning (SFT) to help it reliably adhere to the required format.

\subsection{Online Curriculum Solver}
\label{sec:method:solver}

\begin{algorithm}[t]
\caption{\textbf{\method{}}}
\KwIn{Pretrained base LLM $M_0$; human anchor data $\mathcal{D}_H$; 
batch size $B$; group size $G$; iterations $T$; 
anchor sampling range $k \in [0,5]$; 
curriculum quantiles $[\tau_{\text{low}}, \tau_{\text{high}}]$.}
\KwOut{Trained Challenger $Q_\theta$ and Solver $S_\phi$}

\BlankLine
\For{$t \gets 1$ \KwTo $T$}{
  \tcp{\textbf{Challenger Role: Few-Shot Grounded Generation}}
  \For{$b \gets 1$ \KwTo $B$}{
    Sample $k$ anchor examples: 
    $\mathcal{C}_t \leftarrow \text{Sample}_k(\mathcal{D}_H)$\;
    Generate $G$ question candidates via few-shot learning:
    $\{q_i\}_{i=1}^G \sim Q_\theta(\cdot \mid \mathcal{C}_t, \mathcal{H}_t)$\;
    Evaluate Solver’s success probabilities: 
    $\hat{p}_{\text{succ}}(q_i) \leftarrow \frac{1}{M}\sum_{m=1}^{M} J(q_i, a_i^{(m)})$\;
    
    \eIf{$q_i$ is valid}{
      Compute Challenger reward:
      \begin{equation*}
        r_C(q_i) \leftarrow 
        \bigl(1 - 2|\hat{p}_{\text{succ}}(q_i) - 0.5|\bigr)
        -\lambda_{\text{rep}}\,\mathrm{RepPenalty}(q_t).
      \end{equation*}
    }{
      $r_C(q_i) \leftarrow -\rho_{\text{inv}} \cdot \mathbf{1}[\text{invalid}]$\;
    }
  }
  Update Challenger:
  \[
    \theta \leftarrow \theta + \eta_\theta \nabla_\theta 
    \mathbb{E}_{q_i \sim Q_\theta}[r_C(q_i)] \quad \text{via GRPO.}
  \]
  
  \BlankLine
  \tcp{\textbf{Solver Role: Online Curriculum Learning}}
    Aggregate question–answer pairs $\{(q_i, a_i)\}$ from Challenger and human anchors $\mathcal{D}_H$\;
    Compute success rates $\hat{p}_{\text{succ}}(q)$ for all $q \in \{q_i\} \cup \mathcal{D}_H$\;
    Select mid-difficulty subset:
    $\mathcal{D}_{\text{cur}} \leftarrow 
    \{(q, a) : \tau_{\text{low}} \le \hat{p}_{\text{succ}}(q) \le \tau_{\text{high}}\}$\;
    Set $\mathcal{D}_{\text{mix}} \leftarrow \mathcal{D}_{\text{cur}}$\;
  \For{$(q,a) \in \mathcal{D}_{\text{mix}}$}{
    Compute Solver reward:
    \begin{equation*}
      r_S(q,a) \leftarrow 
      w_{\text{cur}}(q)\mathbf{1}[a = \tilde{y}(q)]
      + \lambda_{\text{hum}} \, w_{\text{hum}}(q)\mathbf{1}[(q,a) \in \mathcal{D}_H]
    \end{equation*}
  }
  Update Solver:
  \[
    \phi \leftarrow \phi + \eta_\phi 
    \nabla_\phi \mathbb{E}_{(q,a) \sim \mathcal{D}_{\text{mix}}}
    [r_S(q,a)] \quad \text{via GRPO.}
  \]
}
\textbf{return} Trained Challenger $Q_\theta$ and Solver $S_\phi$
\end{algorithm}

To ensure efficient learning and avoid collapse, 
the Solver \( S_\phi \) employs an \textbf{online curriculum mechanism} 
that dynamically ranks problems by difficulty and filters the training set to 
the ``zone of proximal learning''. 
At iteration \(t\), the Challenger produces a batch of \(K\) questions 
\(\{q_t^{(i)}\}_{i=1}^K\).
For each \(q_t^{(i)}\), the Solver performs \(M\) rollouts and records the 
success rate \(\widehat{p}_{\text{succ}}(q_t^{(i)})\):
\begin{equation}
    \widehat{p}_{\text{succ}}(q_t^{(i)}) 
    = \frac{1}{M} \sum_{m=1}^{M} J\!\left(q_t^{(i)}, a_t^{(i,m)}\right),
    \qquad
    a_t^{(i,m)} \sim S_\phi(\cdot \mid q_t^{(i)}).
\end{equation}
We then sort all questions (both synthetic and human) by success rate and 
select mid-range items according to the quantile interval 
\([\tau_{\text{low}}, \tau_{\text{high}}]\), empirically we set it as \([0.3, 0.7]\):
\begin{equation}
    \mathcal{D}_{\text{cur}} 
    = \big\{ (q,a) : \tau_{\text{low}} \le \widehat{p}_{\text{succ}}(q) \le \tau_{\text{high}} \big\}.
\end{equation}
This ensures the Solver focuses on problems that are challenging yet solvable,
consistent with optimal learning theory~\citep{Shi2025EfficientRF,Bae2025OnlineDF}.
The Solver reward incorporate both 
synthetic and human data, weighted by curriculum difficulty:
\begin{equation}
\label{eq:rfew-sol}
\mathcal{R}_{\text{sol}}^{\text{\method{}}}(q,a)
= w_{\text{cur}}(q) \cdot \mathbf{1}\{a = \tilde{y}(q)\}
+ \lambda_{\text{hum}} \, w_{\text{hum}}(q) \cdot \mathbf{1}\{(q,a)\in\mathcal{D}_{\text{H}}\},
\end{equation}
where \( w_{\text{cur}}(q) \) is a normalized curriculum weight, and \( w_{\text{hum}}(q) \) upweights the scarce human anchors to prevent forgetting. In practice, we set $\lambda$ as 2.0.
Solver parameters are optimized as:
\begin{equation}
    \phi \leftarrow \phi + \eta_\phi 
    \nabla_\phi \, 
    \mathbb{E}_{(q,a)\sim\mathcal{D}_{\text{cur}}}
    [\mathcal{R}_{\text{sol}}^{\text{\method{}}}(q,a)].
\end{equation}

\subsection{Iterative Co-Evolution}
\label{sec:method:co-evolve}

\textsc{\method{}} maintains the same self-evolving loop as \textsc{R-Zero}, 
but now with a grounded and curriculum-aware data pipeline: Challenger generates synthetic tasks with few-shot demonstrations from human anchors. Solver attempts to solve tasks and estimates success rates. An online curriculum filter selects mid-difficulty tasks from both synthetic and human pools.

\section{Experiments}
\label{sec:experiments}
\subsection{Experiments Setting}
\subsubsection{Baseline Methods}
We employ Qwen3-4B-Base~\citep{Yang2025Qwen3TR} and Qwen3-8B-Base as our backbone models. For comparison, we evaluate against several baselines: 
\vspace{0.02in}
\\
\noindent (1) Base Model — the pretrained checkpoint without any post-training, and the model checkpoint can be downloaded at \url{https://huggingface.co/Qwen/Qwen3-4B-Base} (\url{Qwen3-8B-Base}); 
\vspace{0.02in}
\\
\noindent 
(2) R-Zero~\citep{huang2025r} — the first ungrounded self-play approach where the model generates its own questions and answers, enabling self-improvement through autonomous reasoning; 
\vspace{0.02in}
\\
\noindent 
(3) Absolute Zero~\citep{zhao2025absolute} — a self-play framework constrained to code generation tasks with Python execution as verification, exemplifying domain-specific grounded self-play; 
\vspace{0.02in}
\\
\noindent 
(4) SPICE~\citep{liu2025spice} — a grounded self-play framework that retrieves contexts from a large document corpus to create diverse reasoning tasks with document-grounded answers.

We reproduce baselines (1)-(2) using publicly available implementations under identical training infrastructure, while results for (3)–(4) are taken from the corresponding reports in \cite{liu2025spice}.

We evaluate two variants of our method—\method{} (1\%) and \method{} (5\%)—corresponding to using 1\% or 5\% of examples randomly sampled from the WebInstruct dataset (232k)~\citep{Ma2025GeneralReasonerAL}.

\subsubsection{Evaluation Benchmark}
\label{sec:tasks_and_evaluation}

We assess our framework on a comprehensive suite of benchmarks under two main categories.

\noindent \textbf{Mathematical Reasoning.}
We use five benchmarks: AMC, Minerva~\citep{Lewkowycz2022SolvingQR}, MATH-500~\citep{Hendrycks2021MeasuringMP}, GSM8K~\citep{Cobbe2021TrainingVT} and Olympiad-Bench~\citep{He2024OlympiadBenchAC}.
For evaluation, we follow~\cite{Zhao2025OneTT,Ma2025GeneralReasonerAL} to employ GPT-4o as a programmatic judge to semantically verify the correctness of the final answer against the ground truth. 

\noindent \textbf{General Domain Reasoning.}
To test for the generalization of reasoning ability, we evaluate on the following benchmarks: MMLU-Pro~\citep{Wang2024MMLUProAM}, SuperGPQA~\citep{Team2025SuperGPQASL}, GPQA-Diamond~\citep{rein2024gpqa}, and BBEH~\citep{kazemi2025BIGBenchEH}.
These benchmarks collectively cover a wide range of disciplines (e.g., physics, biology, business, economic, law) with challenging tasks, offering a stricter evaluation of complex reasoning abilities.
For evaluation, we follow \cite{Ma2025GeneralReasonerAL} to report Exact Match (EM) accuracy obtained via greedy decoding.

\subsection{Experimental Results}
\begin{table*}[t]
\centering
\caption{Comprehensive results on reasoning benchmarks. We compare our R-Few against several self-evolving baselines, including R-Zero, Absolute Zero, SPICE. R-Few consistently improves performance over baselines, and approaching General-Reasoner (trained with 232k WebInstruct data) while using substantially less human supervision (1\% and 5\%).}
\vspace{-0.1in}
\scalebox{0.88}{\setlength{\tabcolsep}{1.4mm}{
\begin{tabular}{lccccc|cccc|c}
\toprule
\multirow{3}{*}{\textbf{Models $\downarrow$}} & \multicolumn{5}{c|}{\textbf{Mathmetical Reasoning}} & \multicolumn{4}{c|}{\textbf{General Reasoning}} & \\
\cmidrule{2-11}
& \multirow{2}{*}{AMC} & \multirow{2}{*}{{Minerva}} & {MATH} & \multirow{2}{*}{{GSM8K}} & \multirow{2}{*}{{Olympiad}} & {MMLU} & {Super} & \multirow{2}{*}{{GPQA-D}} & \multirow{2}{*}{{BBEH}} & \multirow{2}{*}{{Avg.}} \\
& & & {500} & & & {Pro} & {GPQA} & & & \\
\midrule
\multicolumn{6}{@{}l|}{\textit{Qwen3-4B-Base}} &&&&& \\
Base Model & 47.5 & 42.3 & 68.2 & 72.6 & 34.8 & 51.6 & 25.4 & 26.3 & 8.1 & 41.9 \\
\quad + R-Zero & 48.2 & 51.2 & 74.8 & 90.6 & 40.6 & 54.2 & 27.8 & 36.4 & 10.4 & 48.2 \\
\quad + Absolute Zero & 50.0 & 41.9 & 76.2 & 89.3 & 41.5 & 52.6 & 27.1 & 35.3 & 8.3 & 46.9 \\
\quad + SPICE & 57.5 & 51.9 & 78.0 & 92.7 & 42.7 & 58.1 & 30.2 & 39.4 & 12.3 & 51.4 \\
\quad + R-Few (1\%) & 52.7 & 52.1 & 77.8 & 92.3 & 42.4 & 55.9 & 29.4 & 35.4 & 11.2 & 49.9 \\
\quad + R-Few (5\%) & 52.4 & 53.2 & 78.0 & 92.6 & 42.8 & 56.2 & 29.4 & 39.9 & 11.8 & 50.7 \\
General-Reasoner & 60.0 & 57.7 & 80.6 & 92.2 & 47.7 & 62.8 & 32.5 & 42.9 & 12.2 & 54.3 \\
\midrule
\multicolumn{6}{@{}l|}{\textit{Qwen3-8B-Base}} &&&&& \\
Base Model & 61.5 & 49.3 & 74.4 & 90.9 & 40.4 & 58.0 & 30.4 & 33.3 & 10.5 & 49.9 \\
\quad + R-Zero & 62.8 & 58.8 & 80.6 & 92.4 & 43.4 & 61.6 & 31.8 & 40.5 & 11.3 & 53.7 \\
\quad + Absolute Zero & 62.5 & 52.9 & 76.6 & 92.0 & 47.8 & 62.5 & 33.5 & 36.8 & 10.8 & 52.8 \\
\quad + SPICE & \underline{70.0} & 59.2 & 79.4 & 92.7 & 42.5 & 65.0 & 35.7 & 39.4 & \textbf{14.9}  & 55.4 \\
\quad + R-Few (1\%) & 69.3 & 59.6 & 81.6 & \textbf{94.0} & 44.0 & 62.8 & 32.7 & 40.4 & 11.8 & 55.1 \\
\quad + R-Few (5\%) & \textbf{72.3} & \underline{60.3} & \underline{82.6} & \underline{93.5} & \textbf{46.4} & \underline{63.2} & \underline{33.5} & \textbf{46.5} & \underline{12.3} & \textbf{56.7} \\
General-Reasoner & 64.8 & \textbf{62.6} & \textbf{83.4} & 92.7 & \underline{46.3} & \textbf{65.1} & \textbf{35.3} & \underline{42.9} & 10.8 & \underline{56.0} \\
\bottomrule
\end{tabular}}}
\label{main_table}
\end{table*}

\subsubsection{Comparison with Baseline Methods}

Table 1 demonstrates that \method{} enables strong and efficient self-evolution across both mathematical and general reasoning tasks. For Qwen3-4B-Base, the Base Model achieves an average score of 41.9, and unguided self-play via R-Zero improves performance to 48.2. However, these gains remain limited, illustrating the difficulty of stable self-evolution without human grounding. In contrast, \method{} produces significantly larger improvements: with only 1\% human data, performance rises to 49.9, and with 5\% data it reaches 50.7, narrowing the gap to General-Reasoner (54.3), which relies on roughly 20× more human-labeled data. This demonstrates that minimal human grounding can unlock meaningful self-evolution while retaining data efficiency.

A notable trend emerges when comparing model scales. The 8B variants exhibit stronger self-evolving capability than their 4B counterparts. While R-Zero provides modest improvements for both, \method{} on Qwen3-8B-Base reaches 55.1 with 1\% data and 56.7 with 5\%, surpassing General-Reasoner (56.0). This indicates that larger models not only benefit more from guided self-evolution but can also nearly replicate or exceed the performance of heavily supervised pipelines with minimal human input.
These observations suggest that as model capacity increases, the model becomes more capable of interpreting human grounding signals, generating higher-quality synthetic data, and improving its reasoning abilities through iterative self-play.

\subsubsection{Ablation Studies}
\label{sec:ablation}

To isolate the contribution of each key component based on our \method{} (5\%) experiments, we conduct a comprehensive ablation study on the \texttt{Qwen3-8B-Base} model. We evaluate the importance of three critical modules by disabling each one individually and measuring its impact on math and general reasoning tasks. The results are summarized in Table~\ref{tab:ablation_study}.

\begin{wraptable}{r}{0.5\textwidth}
    \vspace{-0.17in} 
    \centering
    \small
    \caption{Ablation study on the Qwen3-8B-Base model. Results highlight the impact of challenger training, challenger warm-up, and curriculum learning on downstream reasoning tasks.}
    \vspace{-0.08in}
    \label{tab:ablation_study}
    \begin{tabular}{@{}lcc@{}}
        \toprule
        \textbf{Method} & \textbf{Math} & \textbf{General} \\
        \midrule
        \method{} (from Table \ref{main_table}) & 71.0 & 38.9 \\
        \midrule
        \textit{Ablations} & & \\
        \quad $\vdash$ w/o Challenger Training & 68.0 & 37.4 \\
        \quad $\vdash$ w/o Challenger Warm-up & 69.1 & 37.9 \\
        \quad $\vdash$ w/o Curriculum Learning & 69.1 & 38.1 \\
        \bottomrule
    \end{tabular}
    \vspace{-0.15in}
\end{wraptable}

Among the ablations, disabling challenger training causes the largest degradation, reducing Math and General averages by 1.9 and 1.0 points, respectively. Removing challenger warm-up and curriculum learning yields similar trends, indicating both components are critical for stronger performance. 
Math performance is more sensitive than general domain performance. This sensitivity is consistent with the trend observed in Table~\ref{main_table}, where math benchmarks exhibit larger performance variance across different methods.

\subsubsection{Domain Corelation with Human Data}
\label{sec:analysis_difficulty_accuracy}

\begin{wrapfigure}{r}{0.5\textwidth}
    \centering
    \vspace{-0.4in}
    \includegraphics[width=\linewidth]{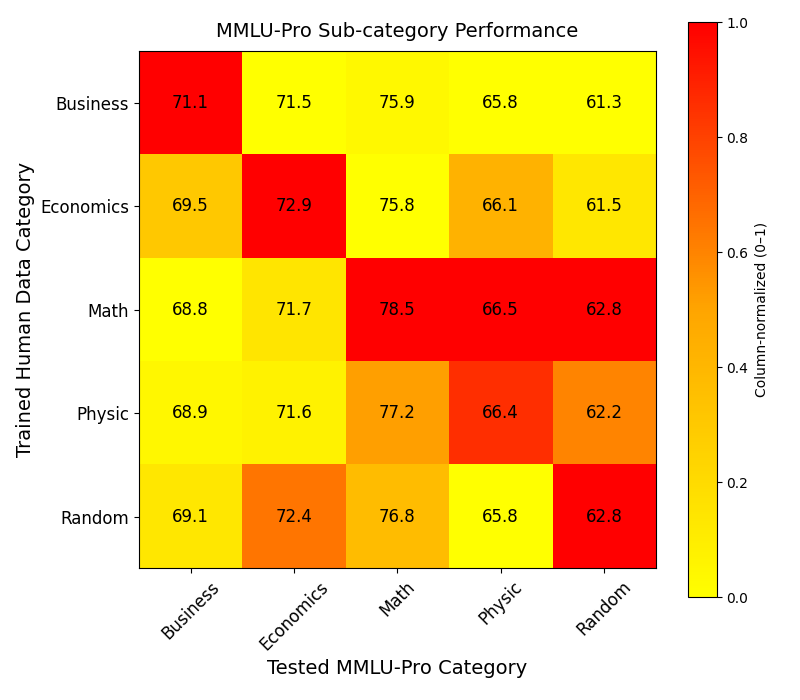}
    \vspace{-0.3in}
    \caption{Impact of domain-sampled human data on performance across MMLU-Pro categories.}
    \vspace{-0.3in}
    \label{fig:domain}
\end{wrapfigure}

Figure \ref{fig:domain} presents the cross-domain performance matrix, showing how human-generated data from each training category influences performance across MMLU-Pro subcategories. 
We sample domain-specific data from WebInstruct (already labeled) to train the self-evolving models, and then evaluate them on MMLU-Pro.
A clear pattern emerges: data from each domain tends to yield the greatest improvements within its own category.
Notably, math stands out as the most useful category, suggesting that mathematical reasoning contributes broadly to complex reasoning capabilities. The cross-domain relationships also highlight two strongly connected pairs: math–physics, reflecting shared quantitative structure, and business–economics, likely because of their overlapping analytical and decision-making frameworks.

Overall, the figure indicates that while heterogeneous human data provides general improvements, domain-specific sampling is particularly effective at strengthening the domain it represents. This observation is valuable because it makes the self-evolving system more controllable: by selecting or emphasizing particular data domains, developers can steer how the model grows, prevent unintended concept drift, and maintain clearer alignment between training inputs and emergent skills, rather than drifting unpredictably into unrelated areas.

\subsubsection{Improved Stability and Reduced Reward Hacking}

\begin{figure}[h]
    \centering   
    \vspace{-0.1in}
    \includegraphics[width=0.325\linewidth]{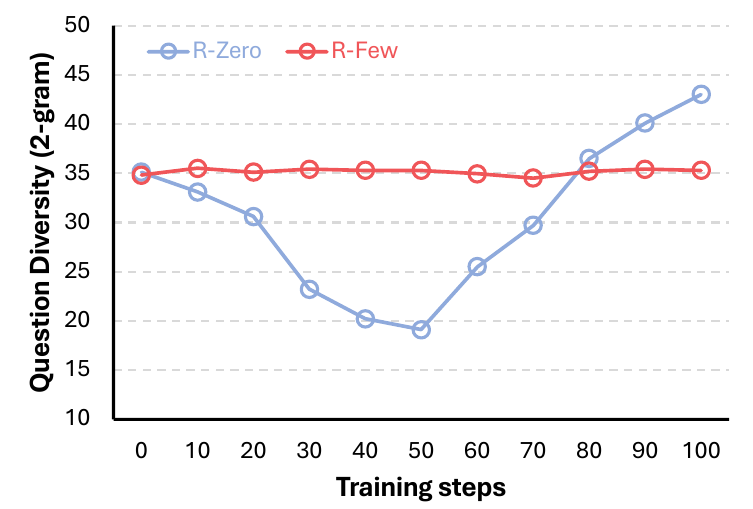}
    \includegraphics[width=0.325\linewidth]{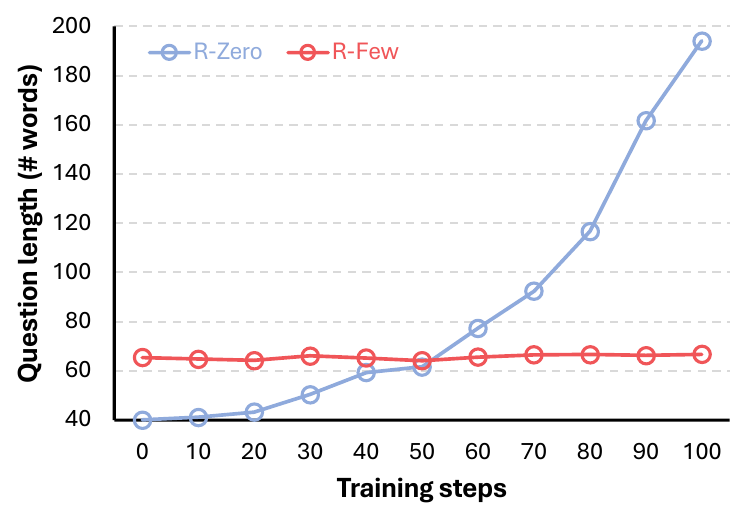}
    \includegraphics[width=0.325\linewidth]{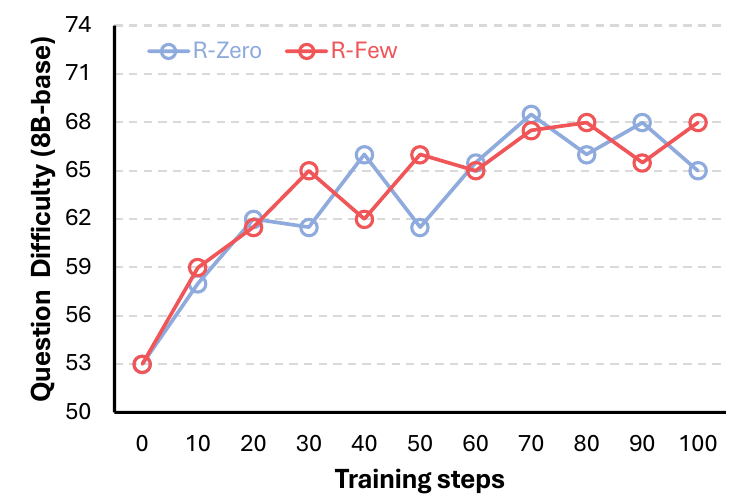}
    \vspace{-0.1in}
    \caption{Training curves of synthetic question diversity (measured by 2-gram lexical diversity), length (measured by word count), and difficulty (evaluated by Qwen3-8B-Base, with ground-truth labeled by Gemini-2.5-Pro) over training. 
    R-Zero collapses in diversity and exhibits length inflation via verbosity, whereas \method{} maintains stable length and diversity during self-evolution.
    }
\end{figure}

As discussed in the introduction, unguided self-evolving systems tend to plateau quickly and exhibit instability, often leading to diversity collapse and reward hacking. 
Our empirical results clearly reflect these phenomena. 
In the left panel, R-Zero shows a sharp decline in question diversity as training progresses, dropping from 35 to below 20 in the first 50 training steps. 
Although the 2-gram diversity score increases after 50 steps, this increase coincides with a dramatic explosion in question length, as shown in the middle panel, indicating that is largely a length-induced artifact rather than genuine semantic diversification. 
This is because longer outputs naturally introduce more unique n-gram combinations, inflating the diversity metric without reflecting meaningful exploration. 
A similar pattern emerges in question length itself. R-Zero progressively inflates question length, eventually producing excessively long questions. This suggests reward hacking, where the model exploits superficial features, such as verbosity, to increase perceived diversity or difficulty signals. 
In contrast, \method{} maintains a stable diversity throughout training while keeping question length consistent, suggesting that lightweight human grounding prevents diversity collapse.

Finally, the right panel shows that both methods increase question difficulty over time, but this similarity masks important qualitative differences. To measure difficulty, we relabel questions generated at different training stages using Gemini-2.5-Pro, then evaluate them by having the same model (Qwen3-8B-Base) attempt to answer them. Difficulty is defined as the percentage of questions the model answers incorrectly (the inverse of accuracy).
We observe that both curves exhibit an overall upward trend, indicating increasing difficulty; however, the source of this increase differs substantially. R-Zero produces questions that are harder to solve primarily because they are longer, rather than because they require deeper reasoning. This reflects another form of reward hacking, where verbosity inflates perceived difficulty without meaningful improvement in reasoning complexity. In contrast, \method{} maintains consistent question length, allowing the model to self-evolve toward genuinely more challenging questions without sacrificing diversity or structure.

\section{Related Work}
\label{sec:related_work}

\subsection{Self-Evolving LLMs and Self-Play}
\label{sec:related_work_self_play}

Early demonstrations of self-play–based learning came from AlphaZero, which learns without external supervision by playing against itself and progressively strengthening its policy through these matches~\citep{silver2017mastering,Silver2017MasteringTG,sukhbaatar2017intrinsic}.
Recent research extended this principle to language leads naturally to self-evolving LLMs that refine their own reasoning capabilities over time~\citep{Tao2024ASO}. 
Several systems instantiate language self-play at scale. Self-play fine-tuning improves relatively weak LLMs by repeatedly generating and solving tasks without large labeled corpora~\citep{chen2024self}. 
This approach has been particularly fruitful in verifiable domains like code generation, where a ``Coder'' agent's program is verified by a ``Tester'' agent's unit tests~\citep{Lin2025LearningTS, Wang2025CoEvolvingLC, Pourcel2025SelfImprovingLM,zhao2025absolute}.
Fully data-free methods such as R-Zero push this further: the model proposes questions, produces candidate solutions, and learns purely from internal verification signals~\citep{huang2025r,chen2025self,kuba2025language}. Other approaches introduce limited seeds or structure, where the model starts from a small set of examples and learns to synthesize increasingly challenging problems around them~\citep{Fang2025SeRLSR,he2025visplay,liu2025spice}.
\method{} builds on this line of work but targets a different point on the supervision spectrum. This guided self-play design aims to preserve the open-ended exploration benefits of self-evolving LLMs while explicitly mitigating concept drift and diversity collapse.

\subsection{Reinforcement Learning for LLM Reasoning}

Reinforcement learning (RL) has become a central mechanism for improving LLM reasoning beyond supervised fine-tuning and instruction tuning, most prominently through reinforcement learning from human feedback (RLHF) and its variants~\citep{kaufmann2024survey}. 
Early RLHF pipelines such as InstructGPT~\citep{ouyang2022training} demonstrated that sequence-level preference optimization can substantially improve helpfulness, safety, and, in many cases, reasoning quality compared to pure supervised fine-tuning. 
Building on these foundations, recent reasoning-centric systems such as DeepSeek-R1~\citep{DeepSeekAI2025DeepSeekR1IR,shao2024deepseekmath} show that RL-style optimization with carefully designed reasoning rewards can significantly enhance multi-step problem solving. Similar R1-style pipelines in both commercial and open-source models further corroborate that tailoring RL objectives to long-form chain-of-thought and verifiable tasks yields substantial gains in reasoning ability~\citep{yang2025qwen3,team2025kimi,liu2025understanding,yu2025dapo}.

Within this landscape, RL with verifiable rewards (RLVR) uses externally defined verifiers or environments to provide scalar feedback~\citep{Ma2025GeneralReasonerAL,su2025crossing,yu2025dapo}. 
For instance, Search-R1 teaches LLMs to reason while leveraging search engines~\citep{Jin2025SearchR1TL}, and semantically-aware R1 variants extend RLVR to open-ended free-form generation~\citep{li2025semanticallyawarerewardsopenendedr1}. These works illustrate how verifiable environments -- logical constraints, executable code, structured prediction targets, or retrieval-enhanced setups -- can provide dense, automatically computable rewards beyond human labels.
Another trend in recent research is label-free reinforcement learning, which aims to improve LLM reasoning without human-annotated data or explicit external verifiers. 
Many such methods use self-generated outputs as a reward signal. This includes leveraging sequence-level confidence~\citep{Li2025ConfidenceIA, Prabhudesai2025MaximizingCA}, the consistency of answers derived from varied reasoning paths~\citep{Zhang2025ConsistentPL,Zuo2025TTRLTR}, or minimizing the output entropy~\citep{Agarwal2025TheUE, Cheng2025ReasoningWE}. These signals are often used within self-training loops where models fine-tune on their own most plausible solutions~\citep{Shafayat2025CanLR, Zhao2025LearningTR}.

\section{Conclusion and Future Work}
\label{sec:conclusion}

In this paper, we introduced \method{}, a guided self-evolving framework that enables LLMs to improve autonomously with minimal human supervision. By combining few-shot–grounded question generation with an online curriculum for solver training, \method{} produces stable, scalable co-evolution and achieves strong gains across mathematical and general reasoning tasks. Despite using only 1–5\% human data, it approaches or surpasses systems trained with far larger labeled datasets, demonstrating the power of lightweight anchoring in preventing drift and enhancing learning dynamics. Future work includes improving efficiency, exploring richer verification methods, and extending self-evolution to open-ended domains lacking objective correctness signals.

\bibliography{ref}
\bibliographystyle{colm}
\newpage
\appendix
\section{Appendix}

\subsection{Training Details}

\begin{wrapfigure}{r}{0.4\textwidth}
    \centering
    \vspace{-0.17in}
    \includegraphics[width=\linewidth]{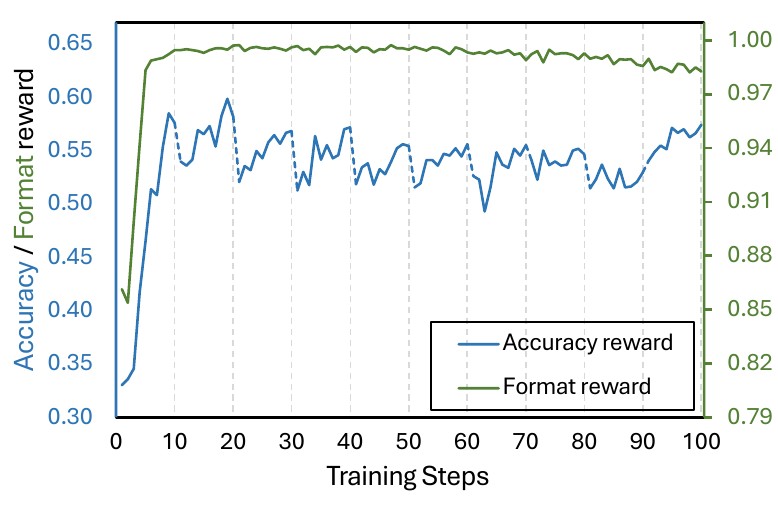}
    \vspace{-0.3in}
    \caption{Training curve of the solver (Qwen3-8B-Base), trained for 100 steps while alternating with the challenger.}
    \label{fig:curve}
    \vspace{-0.1in}
\end{wrapfigure}

Our implementation is based on the EasyR1 \citep{zheng2025easyr1} and R-Zero \citep{huang2025r} codebases.
Figure \ref{fig:curve} shows the solver training curve. The solver is trained for 100 steps, while the challenger is trained for 50 steps. For both models, we use a batch size of 512, a rollout number of 8, and a learning rate of 5e-7.
The challenger and solver are trained iteratively, with the challenger updated for 5 steps followed by 10 steps of solver training. We do not alternate on every step because a single solver update does not significantly change its ability; performing several solver updates per iteration improves training efficiency. At the beginning of each cycle (steps 11, 21, 31, …), the solver is trained on newly generated questions from the challenger. As the challenger generates harder questions, the solver’s accuracy reward drops sharply at each refresh cycle.
For solver training, the reward weights are set to 0.1 for format correctness and 0.9 for accuracy. For challenger training, the uncertainty reward is computed by sampling eight responses from the current solver.

\subsection{Training Hyperparameter}
This section summarizes the  hyperparameters for the Solver and Challenger training stages. All experiments were conducted using BFloat16 (BF16) mixed precision and FlashAttention 2.

\subsubsection{Solver Training}
\begin{itemize}
    \item \textbf{Global Batch Size}: 512
    \item \textbf{Learning Rate}: $5 \times 10^{-7}$
    \item \textbf{KL Penalty Coefficient} ($\lambda_{KL}$): $1 \times 10^{-2}$
    \item \textbf{Max Steps}: 100
    \item \textbf{Number of Rollouts}: 8
    \item \textbf{Rollout Temperature}: 1.0
    \item \textbf{Rollout Top-p}: 0.99
\end{itemize}

\subsubsection{Challenger Training}
\begin{itemize}
    \item \textbf{Global Batch Size}: 512
    \item \textbf{Learning Rate}: $5 \times 10^{-7}$
    \item \textbf{KL Penalty Coefficient} ($\lambda_{KL}$): $1 \times 10^{-2}$
    \item \textbf{Max Steps}: 50
    \item \textbf{Number of Rollouts}: 8
    \item \textbf{Rollout Temperature}: 1.0
    \item \textbf{Rollout Top-p}: 0.99
\end{itemize}

\subsection{Prompt Templates}
This section presents the exact prompt templates used for the solver and challenger models.

All highlighted items marked as ``\texttt{<xxx>}'' will be replaced with the actual input.

\begin{tcolorbox}[
  colback=blue!5!white,
  colframe=blue!75!black,
  title=\textbf{Solver Prompt Template},
  fonttitle=\bfseries,
  sharp corners
]
\textbf{System Message:}
\par 
Please reason step by step, and put your final answer within \textbackslash boxed\{\}.

\textbf{User Message:}
\par
\texttt{<problem\_statement>} \par
\par
\vspace{0.5em}
\end{tcolorbox}

\vspace{1em} 

\begin{tcolorbox}[
  colback=blue!5!white,
  colframe=blue!75!black,
  title=\textbf{Challenger Prompt Template},
  fonttitle=\bfseries,
  sharp corners
]
\textbf{System Message:}
\par
You are an expert problem setter. First, in your private scratchpad, carefully design a brand-new, non-trivial reasoning problem. The subject may be drawn from any high-school or university discipline (mathematics, science, history, literature, social studies, etc.). The problem must be challenging enough that fewer than 30\% of college students would be able to solve it correctly. You will be provided with some example questions, but you are encouraged to brainstorm freely and not be limited by those examples. Finally, output exactly in the following format: \par \vspace{0.5em}
\{question\} \par
\{The complete problem statement on one or more lines\}\\
\{/question\} \par
\par
\vspace{0.5em}
\textbf{User Message:}
\par
\texttt{<Example 1>} \par
\texttt{<Example 2>} \par
\texttt{<Example 3>} \par
\texttt{<Example 4>} \par
\texttt{<Example 5>} \par
Generate a brand-new, challenging reasoning questions now. Each must follow the required format within \{question\} \{/question\}.
\end{tcolorbox}

\subsection{GPT-4o Judge Prompt}
To programmatically evaluate the correctness of answers on mathematical benchmarks where the final answer can be complex (e.g., simplified expressions), we use GPT-4o as a judge. The exact prompt and configuration used for this evaluation are detailed below.

\begin{tcolorbox}[
  colback=blue!5!white,
  colframe=blue!75!black,
  title=\textbf{Configuration for GPT-4o as Judge},
  fonttitle=\bfseries,
  sharp corners
]
\begin{itemize}
    \item \textbf{Model}: \texttt{gpt-4o}
    \item \textbf{Temperature}: 0.0
\end{itemize}

\textbf{System Message:}
\begin{quote}
You are a helpful assistant.
\end{quote}

\textbf{User Message Template:}
\begin{quote}
Look at the following two expressions (answers to a math problem) and judge whether they are equivalent. Only perform trivial simplifications
\par
Examples:\par

    \quad Expression 1: $2x+3$\par
    \quad Expression 2: $3+2x$\par

Yes\par

    \quad Expression 1: 3/2\par
    \quad Expression 2: 1.5\par

Yes\par

    \quad Expression 1: $x^2+2x+1$\par
    \quad Expression 2: $y^2+2y+1$\par

No\par

    \quad Expression 1: $x^2+2x+1$\par
    \quad Expression 2: $(x+1)^2$\par

Yes\par

    \quad Expression 1: 3245/5\par
    \quad Expression 2: 649\par

No\par
(these are actually equal, don't mark them equivalent if you need to do nontrivial simplifications)\par

    \quad Expression 1: 2/(-3)\par
    \quad Expression 2: -2/3\par

Yes\par

(trivial simplifications are allowed)\par

    \quad Expression 1: 72 degrees\par
    \quad Expression 2: 72\par

Yes\par
(give benefit of the doubt to units)\par

    \quad Expression 1: 64\par
    \quad Expression 2: 64 square feet\par

Yes\par
(give benefit of the doubt to units)\par

---\par
\end{quote}
\end{tcolorbox}

\begin{tcolorbox}[
  colback=blue!5!white,
  fonttitle=\bfseries,
  sharp corners
]
\begin{quote}
YOUR TASK\par

Respond with only "Yes" or "No" (without quotes). Do not include a rationale.\par

    Expression 1: \texttt{<Expression 1>}\par
    Expression 2: \texttt{<Expression 2>}\par
\end{quote}
\end{tcolorbox}



\end{document}